\documentclass{article}

\usepackage{microtype}
\usepackage{graphicx}
\graphicspath{./images/}
\usepackage{amsmath}
\DeclareMathOperator*{\argmax}{arg\,max}

\usepackage{subfigure}
\usepackage{booktabs} 

\usepackage{hyperref}

\usepackage[accepted]{icml2021}

\icmltitlerunning{Interpretability and Explainability in NLP}

\begin{document}

\abovedisplayskip=0pt
\abovedisplayshortskip=-0pt
\belowdisplayskip=0pt
\belowdisplayshortskip=0pt

\twocolumn[
\icmltitle{Quantifying Explainability in NLP and Analyzing Algorithms for Performance-Explainability Tradeoff}



\icmlsetsymbol{equal}{*}

\begin{icmlauthorlist}
\icmlauthor{Mitchell Naylor}{sm}
\icmlauthor{Christi French}{sm}
\icmlauthor{Samantha Terker}{sm}
\icmlauthor{Uday Kamath}{sm}
\end{icmlauthorlist}

\icmlaffiliation{sm}{Digital Reasoning / Smarsh, Inc}

\icmlcorrespondingauthor{Mitchell Naylor}{mitch.naylor@digitalreasoning.com}
\icmlcorrespondingauthor{Uday Kamath}{uday.kamath@digitalreasoning.com}

\icmlkeywords{Machine Learning, ICML, Interpretable Machine Learning, Explainable Machine Learning, NLP, BERT, Text Classification, Explanation Quality}

\vskip 0.3in
]




\begin{abstract}

The healthcare domain is one of the most exciting application areas for machine learning, but a lack of model transparency contributes to a lag in adoption within the industry. In this work, we explore the current art of explainability and interpretability within a case study in clinical text classification, using a task of mortality prediction within MIMIC-III clinical notes. We demonstrate various visualization techniques for fully interpretable methods as well as model-agnostic \emph{post hoc} attributions, and we provide a generalized method for evaluating the quality of explanations using infidelity and local Lipschitz across model types from logistic regression to BERT variants. With these metrics, we introduce a framework through which practitioners and researchers can assess the frontier between a model's predictive performance and the quality of its available explanations. We make our code available\footnote{Code available at \url{https://github.com/mnaylor5/quantifying-explainability}\label{supfoot}} to encourage continued refinement of these methods.

\end{abstract}

\printAffiliationsAndNotice{}  

\section{Introduction}
\label{intro}

Healthcare is an extremely promising area for applying machine learning solutions -- clinical support systems can aid physicians with decision making, text-based solutions can help match patients to relevant clinical trials, and ML systems can save lives by identifying patterns indicative of sepsis, stroke, or cancer. However, the healthcare domain also presents immense legal and ethical considerations, and model mistakes can cost patient lives. Clinical stakeholders not only need to know that models have been faithfully developed, but they also desire transparency in the output of the model so that they can more easily diagnose model failures. A lack of explainability in healthcare, especially clinical decision support models, has been criticized \cite{10.1001/jama.2018.17163}, leading to additional innovation and progress in the development of explainable solutions \cite{lauritsen2019explainable}.

Historically, a trade-off between model performance and explainability has been described, and lower performing yet explainable models are often preferred to more sophisticated black box models due to lack of trust or the presence of strict transparency requirements (e.g. regulatory constraints). To address this, model-agnostic \emph{post hoc} explainability methods such as LIME \cite{ribeiro2016why} and SHAP \cite{lundberg2017unified} were developed, where a second model is trained to explain the original predictions. However, there have been issues reported with these methods: they do not always faithfully describe the original model's behavior and can give different explanations for similar examples \cite{alvarezmelis2018robustness}. Recent efforts, namely \cite{article}, call for the use of inherently interpretable models in situations where human understanding is necessary -- especially when consequences from undue faith in model output can be dire. These methods generally produce sparse models which allow human stakeholders to gain understanding about the broader system in which the model operates, as well as understand precisely how the model produces each prediction.

Natural language processing (NLP) presents a unique modeling and interpretability challenge. Many traditional methods of text preparation are transparent in nature, but as Transformer-based models \cite{vaswani2017attention} have gained popularity, we continue to see the boundaries of black-box modeling pushed. This raises a pressing issue for data scientists and their stakeholders within the healthcare domain: how should we appropriately balance the need for well-performing models with the desire for human interpretability of models and their outputs? 

With such a wide variety of available tools, it should also be noted that the use of these techniques should be tailored to the audience -- for example, data scientists may be interested in interactively exploring neuron activations, while a physician may be interested in the features present in a medical record that contributed to a clinical decision.

\subsection{Interpretability versus Explainability}
In Rudin's work \citeyearpar{article}, she defines ``interpretable" machine learning as the use of models which themselves are fully interpretable by human users, as opposed to using ``explainability" techniques to understand the behavior of non-interpretable models. We follow these definitions throughout, although we acknowledge that there is some degree of gray area with certain attribution methods.

Importantly, neither of these concepts are in any way causal -- all of the methods in this paper should be used to gain insight into how a specific model works, rather than providing any understanding of underlying causal structures of the broader system in which the models operate.

\subsection{Contributions}
The major contributions of this work are as follows:

\begin{itemize} 
    \itemsep=-.2em
    \item Quantifying the quality of available explanation methods using infidelity \cite{yeh2019infidelity}, generalizing perturbation methods for local Lipschitz \cite{alvarezmelis2018robustness} on text classification, and making these methods available for use in open-source form
    \item Comparing the performance of traditional algorithms with interpretable ML techniques and state-of-the-art methods in a practical case study in clinical text classification
    \item Reaching state-of-the-art performance on the mortality prediction task (among models without additional MIMIC-specific pretraining) using BigBird \cite{zaheer2021big}
    \item Discussing the distinction between \emph{interpretability} and \emph{explainability}, practical issues in measuring and implementing various explainable and interpretable algorithms, and particularly the challenges facing practitioners within the healthcare industry
\end{itemize}

\section{Experiment Setup}
\label{setup}

\subsection{Data}
Given the sensitive nature of healthcare data, few openly accessible datasets of clinical text exist for broad research use. In collaboration with Beth Israel Deaconess Medical Center, Johnson et al. introduced MIMIC-III \citeyearpar{Johnson2016} -- a database containing deidentified electronic medical record (EMR) output of Intensive Care Unit (ICU) stays, including structured data elements of medication and treatment as well as unstructured clinical text. Because MIMIC-III is the only publicly available data source of its type and scale, it frequently serves as the basis for peer reviewed studies which make use of EMR data. Here, we utilize the unstructured discharge summaries for our classification task.

\subsection{Classification Task}
We selected one task in order to compare interpretable features across model types and explainability methods within the context of a practically relevant use-case. We use the single-label, binary classification of in-hospital mortality prediction from simulated admissions notes. This benchmark was established in Van Aken et al. \citeyearpar{van-aken-etal-2021-clinical}, and we follow their process of data preparation. Details about this task dataset are provided in supplementary material\textsuperscript{\ref{supfoot}}. 

\section{Methods}
\subsection{Classifiers}
We include several classifiers with varying levels of complexity and interpretability: 

\begin{itemize}
    \itemsep=-.2em
    \item Logistic Regression (LR) 
    \item Random Forest (RF) \cite{breiman2001random}
    \item Explainable Boosting Machine (EBM) \cite{nori2019interpretml}
    \item DL8.5 \cite{Aglin_Nijssen_Schaus_2020}
    \item Boosted Rule Sets \cite{FREUND1997119}
    \item Bayesian Rule Lists \cite{yang2017scalable}
    \item Optimal Classification Trees \cite{10.1007/s10994-017-5633-9}
    \item CORELS \cite{angelino2018learning}
    \item BigBird \cite{zaheer2021big}
\end{itemize}

\subsection{Explainability}

For our traditional models (LR, RF, EBM), we selected two popular \emph{post hoc} explainability methods: Local Interpretable Model-agnostic Explanations (LIME) \cite{ribeiro2016why} and SHapley Additive exPlanations (SHAP) \cite{lundberg2017unified}. Since some of our models are also inherently interpretable (LR and EBM), we were also able to compare \emph{post hoc} explanations to the true local explanations.

Transformer models are a rapidly evolving area of active research, and efforts are ongoing to develop explainability techniques for better understanding their underlying processes. Input attribution methods show importance of input tokens, illustrating the degree to which each token contributed to a given prediction. Several algorithms exist for performing input attribution with deep learning models: saliency \cite{simonyan2014deep} and integrated gradients \cite{sundararajan2017axiomatic} are somewhat agnostic methods which use various backpropagation features to perform attribution, while other methods exist for more specific purposes (e.g. deconvolution and GradCAM for computer vision). In addition to these methods, SHAP and LIME are compatible for \emph{post hoc} explainability with deep learning models -- however, both have been shown to possess worse qualities than techniques with properties native to deep learning \cite{alvarezmelis2018robustness}. 

Along with the growing trend of Transformer interpretability features, researchers have created a number of visualization techniques to highlight various behaviors (self-attention weights, activations, etc.). BertViz \cite{vig-2019-multiscale} and Ecco\footnote{Accessed via \url{https://github.com/jalammar/ecco}} \cite{alammar2020explaining} are two libraries which provide useful front-ends for visualizing such behaviors within Transformer models, and LIT \cite{tenney2020language} is a more general framework for practitioners to visualize and perform behavioral testing on various NLP models. 

\subsection{Explanation Quality Metrics}
With a variety of options available for input attribution and local explainability for ``black-box" models, a user will naturally wonder which method best represents the underlying mechanisms of the model. There are a handful of metrics which seek to quantify the quality of model explanations \cite{electronics10050593}, however there is no single metric which will adequately cover all aspects of interpretability and apply to all contexts. For example, parsimony of the global explanations (i.e. the inherent human interpretability of a model's structure) may be more important when developing clinical guidelines, while fidelity of prediction explanations might adequately capture the needs of a human-in-the-loop ML system. 

For our case study, we searched for metrics which measure various components contributing to the quality of \emph{post hoc} explanations provided through input attribution methods (e.g. SHAP, integrated gradients, etc.). 

\subsubsection{Generalizing Local Lipschitz}
Local Lipschitz \cite{alvarezmelis2018robustness} is an explanation quality metric which measures the relative change in attributions introduced by relatively small perturbations in the model input. This metric is closely related to the notion of a particular explanation method's sensitivity when slight changes are introduced -- if the input changes by some small amount, then we would presume that the explanations should not change drastically. The formula for local Lipschitz is given as follows:
$$ \hat{L}(x_i) = \argmax_{x_j \in B_\epsilon(x_i)} 
\frac{||f(x_i) - f(x_j)||_2}{||x_i - x_j||_2}$$

One challenge came to light when applying this metric to text data: the concept of a \emph{local neighborhood} is difficult to define within the context of raw text -- we can measure the distance between vectorized documents through metrics such as $\ell2$ distance or cosine similarity, but there is no consistent radius ($\epsilon$ in the formula) which can be equivalently applied to models with different preparation types (e.g. TF-IDF document vectors and a sequence of word embeddings). We address this issue by introducing a text perturbation method based upon token replacement via nearest neighbor sampling. Pseudocode for our document perturbation method is given in Algorithm ~\ref{alg:perturbation}. Each token -- a single word in a document to be vectorized, or a single token ID to be fed into a language model -- has probability $\pi$ of being selected for replacement, and each selected token is replaced by sampling from its $k$ nearest neighbors in vector space. -- in this description, \verb+sample_neighbors+ is simply shorthand for sampling a replacement token from its closest neighbors. For traditional methods we use a word2vec model \cite{rehurek_lrec, mikolov2013efficient} trained on the mortality training set. For the BigBird model, we construct a distance matrix of tokens using pairwise $\ell2$ distance between entries within the language model's word embedding layer.

\begin{algorithm}[tb]
  \caption{Text Perturbation}
  \label{alg:perturbation}
\begin{algorithmic}
  \STATE {\bfseries Input:} document $Doc$, sampling probability $\pi\in [0, 1]$, number of nearest neighbors to consider $k>0$
  \STATE {\bfseries Output:} modified document $modifiedDoc$, with tokens randomly replaced
  \STATE Initialize $modifiedDoc = [\ ]$.
  \FOR{$token$ {\bfseries in} $Doc$}
  \IF{$random.uniform() \leq \pi$}
  \STATE $replacement = \operatorname{sample\_neighbors}(token, k)$
  \STATE $modifiedDoc.append(replacement)$
  \ELSE
  \STATE $modifiedDoc.append(token)$
  \ENDIF
  \ENDFOR
\end{algorithmic}
\end{algorithm}

Once our perturbation method is defined, we can construct a neighborhood of perturbed documents for each document in our sample for local Lipschitz estimation. For our experiments, we used a token sampling probability of $\pi=0.1$ and sampled from the nearest $k=10$ neighbors. For each point in the neighborhood, we calculate new attributions based on the perturbed input -- the $\ell2$ distance between perturbed and original attributions serves as the numerator of the local Lipschitz ratio, and the denominator is given by the $\ell2$ distance between perturbed and original numericalized inputs (i.e., embedded sequence or vectorized document after perturbation). In order to enforce the neighborhood aspect of local Lipschitz, we apply a radius of $\epsilon=0.25$ for removing perturbed inputs which lie too far outside the local neighborhood. 

\subsubsection{Infidelity}
Infidelity \cite{yeh2019infidelity} is another metric which seeks to measure the degree to which the explanations capture the change in true model output when ``significant" perturbations are introduced. 

Infidelity is somewhat less exposed to the problem of neighborhood definition than local Lipschitz because it has no need for a common radius. It does, however, require that the perturbations introduced can be subtracted from the input, thus requiring that the perturbations and model inputs are both represented numerically. This is a less desirable aspect of infidelity as a metric to compare across modeling paradigms in NLP, so we account for this by using Gaussian noise with zero-mean and standard deviation proportional to the relevant preparation method (i.e., embedded tokens for LMs and TF-IDF vectors for traditional models). 

\section{Experimental Results}
\subsection{Classifier Performance}
In keeping with the benchmark task \cite{van-aken-etal-2021-clinical}, models are evaluated using the mortality prediction test set and the AUC metric. While we built models with the intention of optimizing validation performance, reaching state-of-the-art test set performance on the benchmark leaderboard was not our primary goal -- our intention is to demonstrate various interpretability methods using a variety of realistic models, so we trained until reaching parity with the performance. For each model, test set performance is provided in Table~\ref{model-performance} along with its most comparable benchmark.

Transformer-based models outperformed both traditional methods and interpretable models. BigBird performed best, beating BioBERT \cite{DBLP:journals/corr/abs-1901-08746} and performing comparably to the benchmark models which were pretrained on additional MIMIC data: CORe and DischargeBERT \cite{alsentzer2019publicly} (also known as \emph{BioClinical BERT}). 

Of the inherently interpretable methods, LR and EBM performed comparably to, if slightly better than, the established ``BOW" benchmark, while DL8.5 and Boosted Rule Sets performed significantly worse than any reasonable baseline. The fact that a decision tree performed poorly on a large-scale NLP task is somewhat unsurprising -- however, the DL8.5 algorithm given one hour to train failed to reach competitiveness with a baseline CART model with all else held constant (tree depth, feature representation, etc.). The final traditional model we used with success is random forest, which also slightly outperforms the ``BOW" benchmark AUC.

\subsubsection{Computational Challenges with Interpretable Methods}
In our experiments, we saw that many of the available interpretable and/or provably optimal techniques are too computationally complex to return the provably optimal solution within a reasonable time on the document-term matrices for our task. Under the category of optimal trees, we unsuccessfully attempted the methods introduced in \cite{10.1007/s10994-017-5633-9} and implemented in PyOptree\footnote{Accessed via \url{https://github.com/pan5431333/pyoptree}} -- this took several hours to complete single iterations on a single tree model. For rule lists, we attempted Bayesian Rule Lists \cite{yang2017scalable} and CORELS \cite{angelino2018learning}, neither of which could work due to the scale of the data for this task.

\begin{table}[t]
\caption{Classifier performance (AUC) on the mortality prediction test set along with most comparable benchmarks.
\newline---\textsuperscript{*}: Unable to execute within a reasonable time or compute limit.}
\label{model-performance}
\vskip 0.15in
\begin{center}
\begin{small}
\begin{sc}
\begin{tabular}{lcp{3cm}}
\toprule 
Model & AUC & Benchmark AUC \\
\midrule
LR & 81.07 & 79.15 \\
\hline
RF       & 79.30 & 79.15 \\
\hline
EBM                 & 76.50 & 79.15 \\
\hline
DL8.5               & 51.02 & 79.15 \\
\hline
Boosted   & 73.49 & 79.15 \\
Rule Sets & & \\
\hline
BigBird             & 83.59 & 81.13 (BERT) \newline 82.55 (BioBERT) \newline 84.51 (\emph{DischargeBert}) \\
\hline
PyOptree       & ---\textsuperscript{*} & 79.15 \\
\hline
CORELS       & ---\textsuperscript{*} & 79.15 \\
\hline
Bayesian       & ---\textsuperscript{*} & 79.15 \\
Rule Lists & & \\
\bottomrule
\end{tabular}
\end{sc}
\end{small}
\end{center}
\vskip -0.1in
\end{table}

\subsection{Demonstrating Interpretability}
Input attribution is an obvious choice for visualizing the local explanations produced by both high-dimensional interpretable models (such as LR and EBM) and deep learning models. Such visualizations can be understood by less technical human users -- including clinicians -- and can easily be integrated into a workflow in which the users interact with the information being classified. A well designed attribution visualization will require minimal education or training for human stakeholders who are familiar with the model's task. In a text-based setting, this can be achieved simply by highlighting the terms within the document which contributed to a certain prediction. This level of information can help bolster trust in the model, while also helping clinical experts identify potential root causes of prediction errors. 

We use input attribution for local explainability throughout our experiments -- while we focus on text classification for this work, this same concept is perfectly compatible with other NLP tasks, including named entity recognition (NER) and language modeling, and many of the libraries we use can be mapped directly to other tasks within and beyond NLP. 

\subsubsection{Comparing Surrogate Attributions to True Local Explanations}
The \verb+interpret+ library \cite{nori2019interpretml} allowed us to directly compare the local explanations obtained by LIME and SHAP to the underlying truth obtained by enumerating the effects from the learned models for logistic regression and EBM.

\begin{figure}[!htb]
\centering
    \begin{subfigure}[True local explanations for logistic regression.]{
        \centering 
        \includegraphics[width=0.5\textwidth]{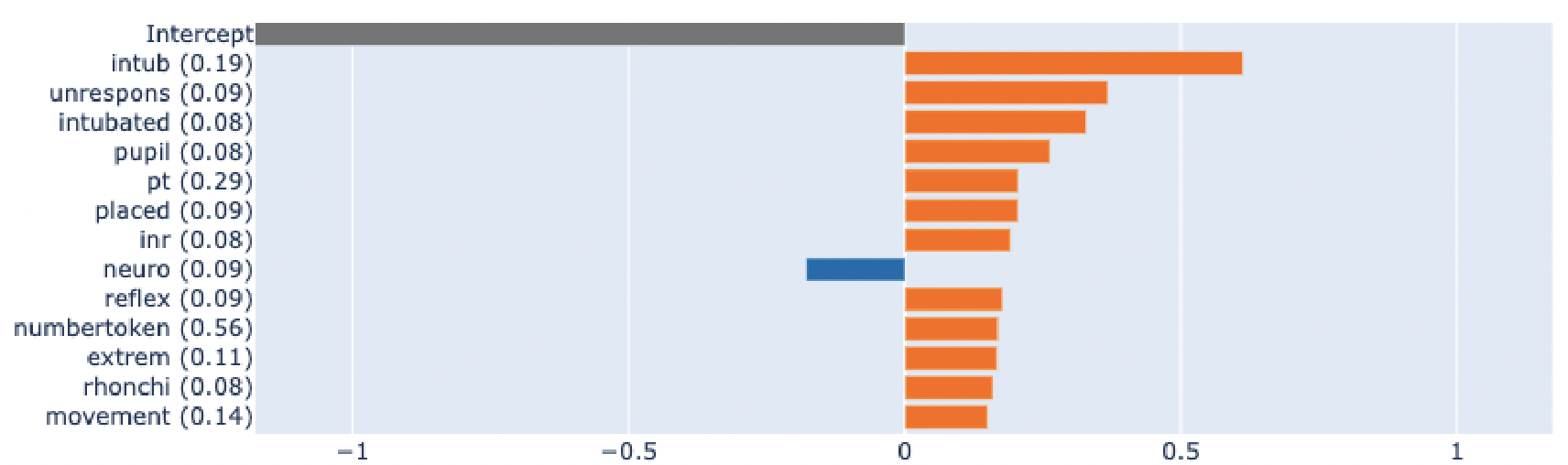}
        \label{fig:lr-interpret-truth}}
    \end{subfigure}%
    \qquad
    \begin{subfigure}[SHAP-based explanations.]{
        \centering
        \includegraphics[width=0.5\textwidth]{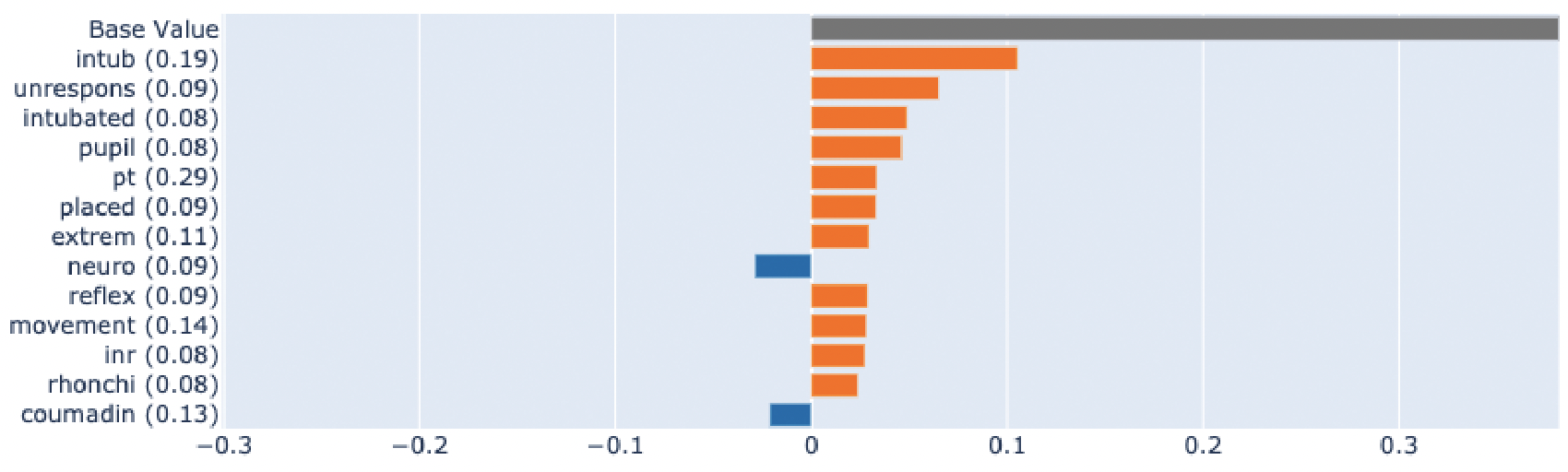}
        \label{fig:lr-interpret-shap}}
    \end{subfigure}
    \qquad
    \begin{subfigure}[LIME-based explanations.]{
        \centering
        \includegraphics[width=0.5\textwidth]{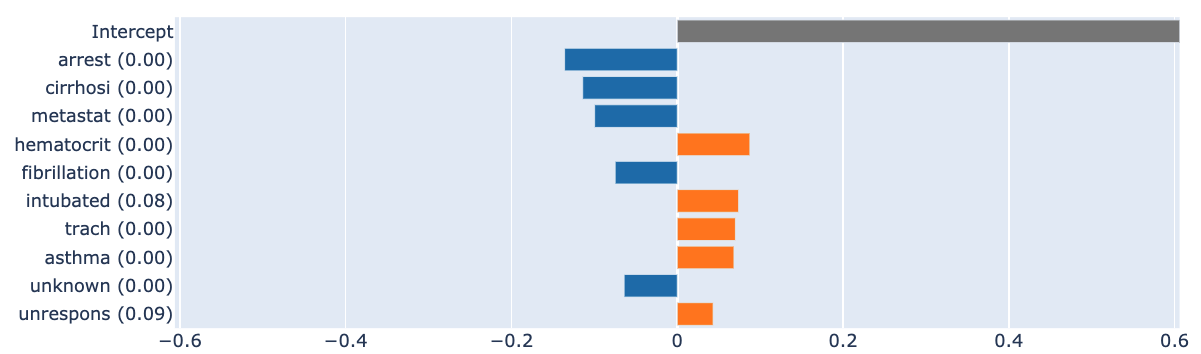}
        \label{fig:lr-interpret-lime}}
    \end{subfigure}
    \caption{Comparison of true model features to \emph{post hoc} feature explanations on a single observation.}\label{fig:lr-interpret}
    \vskip -0.1in
\end{figure}

Figure \ref{fig:lr-interpret} shows a comparison of actual local explanations from a logistic regression versus results from \emph{post hoc} explainability methods SHAP and LIME. SHAP features match actual LR features almost exactly, with only two exceptions: ``numbertoken" (our catch-all token for numeric values in the text) was found in LR but not SHAP, while ``coumadin" was found in SHAP but not LR. LIME features had very little overlap with the true LR features; only two features -- ``intubated" and ``unrespons" -- were found in both. This is unsurprising, as LIME has been reported to have low robustness \cite{alvarezmelis2018robustness}. EBM explainability follows a similar pattern where SHAP explains the feature importance more accurately than LIME, and additional comparisons can be found in our supplemental material. Between the two classification methods, there was also some feature overlap, such as ``intubated,"  ``neuro," and ``coumadin."

\subsubsection{Global Interpretability}
\begin{figure}[!htb]
\centering
    \begin{subfigure}[Global feature importance for EBM]{
        \centering 
        \includegraphics[width=0.5\textwidth]{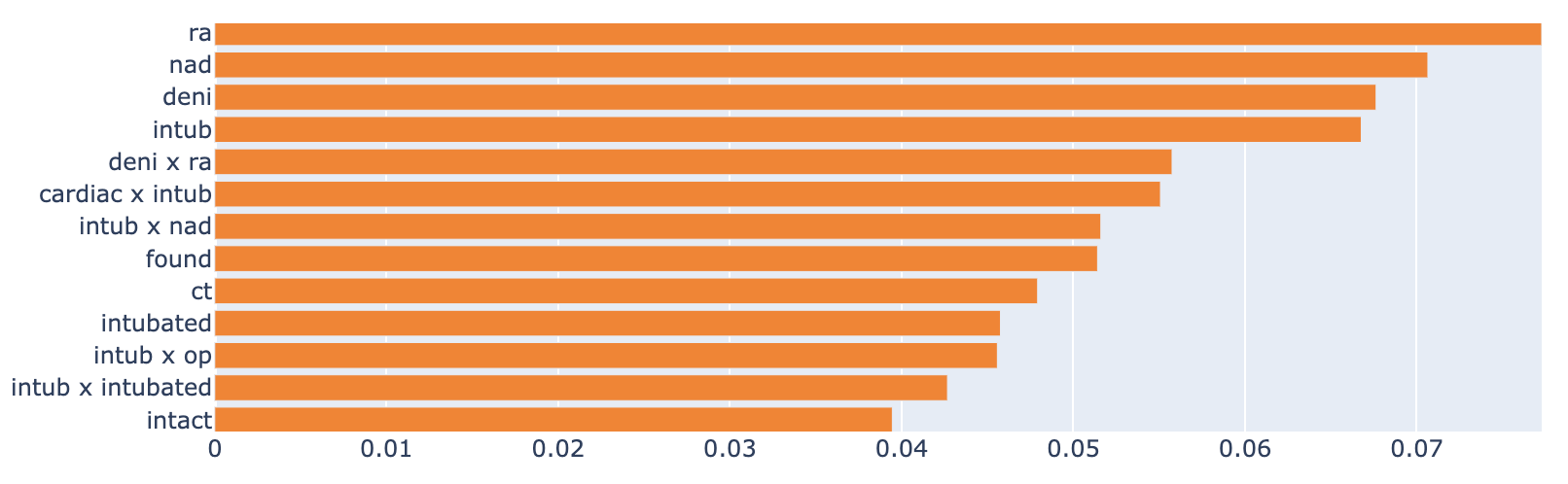}
        \label{fig:ebm-global-importance}}
    \end{subfigure}%
    \qquad
    \begin{subfigure}[Impact of TF-IDF score of ``intub"]{
        \centering
        \includegraphics[width=0.5\textwidth]{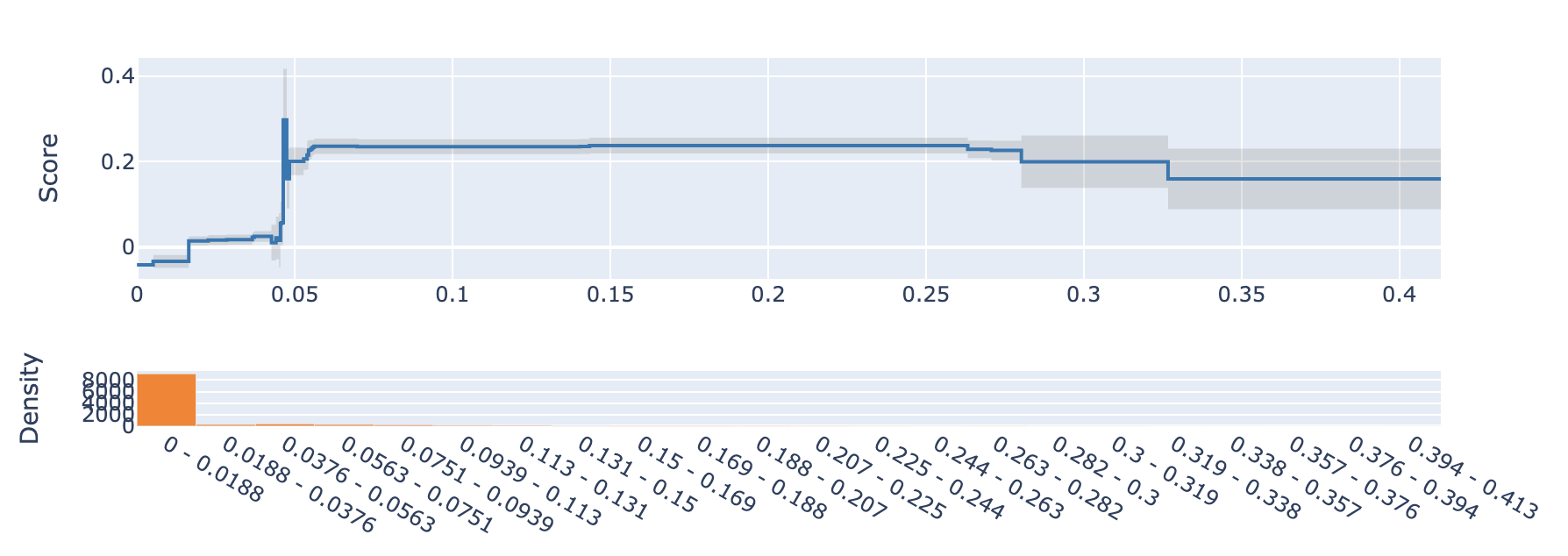}
        \label{fig:ebm-intub}}
    \end{subfigure}
    \caption{Global interpretability for an EBM model from the interpret library.}
    \label{fig:ebm-global}
\end{figure}

Fully interpretable models generally include a method for global interpretability -- a set of coefficients, tree structure, or something similar indicating the ``important" features or an explicit decision path detailing the model's behavior under any conceivable permutation. Various options exist for displaying feature importance -- for example, a tree-based model can be plotted in graph form, raw feature importance for LR/EBM can be displayed similarly to the local explanations in the previous section, and individual feature effects can be plotted over the range of feature values as in a partial dependence plot (PDP). Figure \ref{fig:ebm-global} illustrates the global interpretability features provided by the ``glassbox" models in the \verb+interpret+ library: global feature importance for the EBM model can be seen in \ref{fig:ebm-global-importance}, and \ref{fig:ebm-intub} shows the individual modeled impact of the feature ``intub".

\subsubsection{Interpretable Classifiers}
We chose two interpretable methods which are not commonly used in NLP problems: Boosted Rule Sets \cite{FREUND1997119} and DL8.5 \cite{Aglin_Nijssen_Schaus_2020}. These methods bring something slightly different to the conversation around interpretability than EBM and logistic regression, in that they produce models which are more sparse in nature and can easily be understood by a human user. In the case of DL8.5, there is the additional benefit of the final model being provably optimal for a given specification and dataset.

We used the implementation of Boosted Rule Sets in the \verb+imodels+ library \cite{Singh2021}, and our best model achieved test AUC of 73.49 -- considerably worse than the benchmark performance. This model learned an ensemble of 50 simple rules which indicate higher likelihood for in-hospital mortality, in the format of $token \leq score$, where $token$ is an n-gram found in the document term matrix, and $score$ is the TF-IDF score for that n-gram in the given document. A subset of these rules is included in supplementary material.

DL8.5 performed extremely poorly on this task due to the dimensionality of the dataset. Because of the length and diverse content of the admission notes, we retain as much of the information as possible through our document preprocessing. The high dimensionality of the design matrix -- in this case, both cardinality and volume -- overwhelms the search algorithm, which essentially cannot run without time limits. When this time limit is reached, a poor model is returned, achieving virtually no AUC lift. DL8.5 and Boosted Rule Sets would benefit from additional feature engineering; however, lexicons and hand-engineered features are expensive to develop and maintain, and other dimensionality reduction techniques (e.g. manifold learning for high-dimensional feature spaces) would sacrifice the level of interpretability. Both of these alternative approaches are beyond the scope of this work.

\subsubsection{Transformer Explainability: Input Attribution and Behavioral Visualization}
\begin{figure*}
    \centering
    \includegraphics[width=0.7\textwidth]{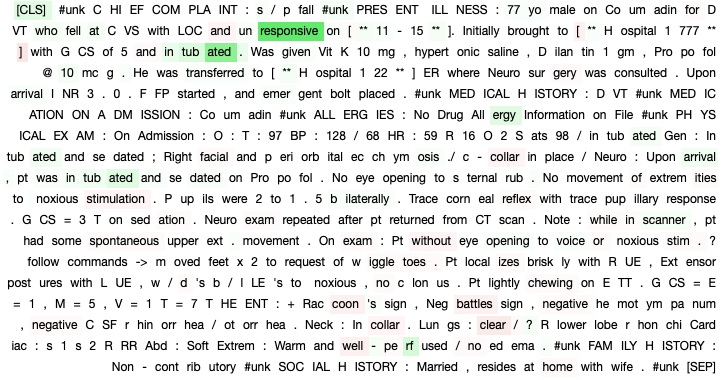}
    \caption{Saliency-based local explanations from the Captum library.}
    \label{fig:captum-attribution}
\end{figure*}

We explored a variety of visualization techniques for Transformer explainability, including input attribution and visualization of activations and attentional weights. The attribution methods produce results which are conceptually similar to those given by SHAP and LIME: they are designed to denote the degree to which a specific input token contributed to the model's output -- in this case, mortality prediction. Figure \ref{fig:captum-attribution} shows input attribution calculated via the saliency method, and visualized using the Captum library \cite{kokhlikyan2020captum} for a true positive prediction (a patient correctly predicted to expire while in the ICU). We can see that there are tokens such as ``unresponsive" and ``intubated" highlighted green, indicating a positive influence (pushing the model toward predicting patient expiration). Additionally, some tokens have a red highlight indicating an impact that suppresses predicted mortality, including ``clear" within a statement describing the patient's lungs. 

Beyond input attribution, other methods are tailored to gaining deeper understanding of the nuances within the representations learned by the language model. Alammar's Ecco library provides an interactive visualization to highlight the input tokens which produce similar neuron activations. Ecco uses non-negative matrix factorization to reduce the high-dimensional activation data into a user-specified number of factors, which are then displayed interactively.

\begin{figure}[htb]
\centering
\includegraphics[width=0.5\textwidth]{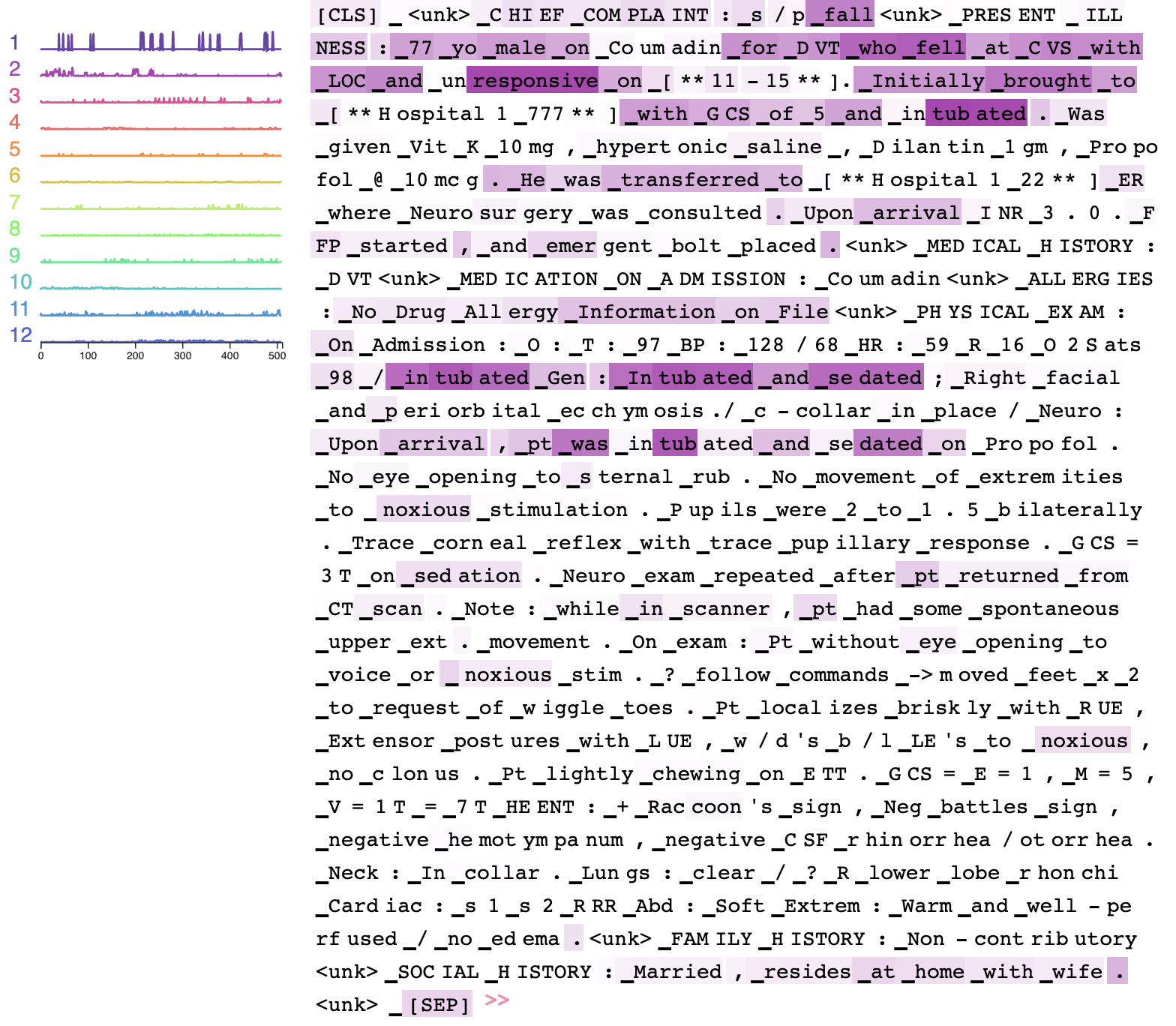}
\caption{Ecco's non-negative matrix factorization of neuron activations with component 2 highlighted.}\label{fig:ecco}
\end{figure}

Figure \ref{fig:ecco} shows Ecco's results for the same true positive case: neuron activations are reduced to 12 factors, and related tokens are highlighted. The sparklines on the left side of the image correspond to individual factors, along with their activations throughout the document. In this view, Factor 2 is highlighted alone, where we can see that it includes language relation to intubation, sedation, falling, and unresponsiveness. This is helpful for us to glean insights from potentially related terms. Additional views can be found in our supplemental material -- for example, in addition to this factor, switching to Factor 11 seems to highlight specific mentions of medication (Coumadin, Dilantin, Propofol) as well as generic treatments (hypertonic saline, Vitamin K).

\begin{figure}[h!]
\centering
    \begin{subfigure}[Full attention weights displayed.]{
        \centering 
        \includegraphics[width=0.3\textwidth]{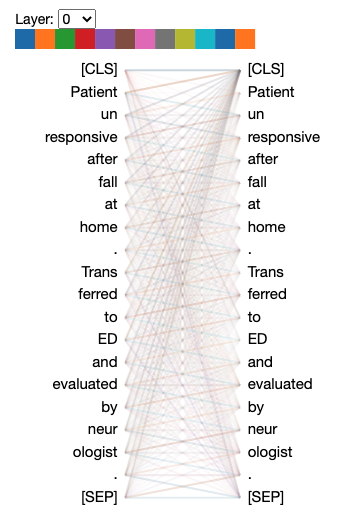}
        \label{fig:bertviz-full}}
    \end{subfigure}%
    \begin{subfigure}[Inbound attention weights for a subword of ``unresponsive" isolated.]{
        \centering
        \includegraphics[width=0.3\textwidth]{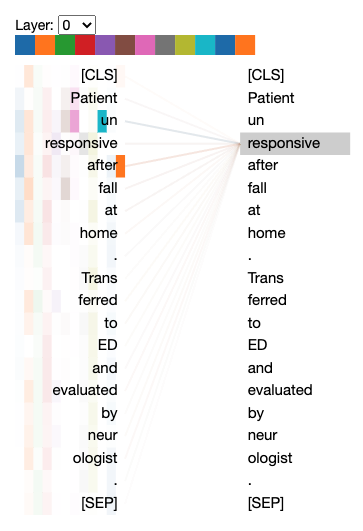}
        \label{fig:bertviz-isolated}}
    \end{subfigure}
    \vskip -0.05in
    \caption{Attention weight visualization with BertViz.}\label{fig:bertviz}
    \vskip -0.15in
\end{figure}

We also explored libraries for visualizing full self-attention weights. BertViz \cite{vig-2019-multiscale} has nice built-in interactivity to allow users to explore results by layer and attention head. However, one drawback of this method is that it does not perform particularly well with relatively long input sequences: the task of visualizing full self-attention across 1,000 tokens is a difficult task, and existing tooling struggles to accommodate such sequences. We can instead use these techniques to visualize local effects by viewing self-attention weights within a smaller segment of a larger document. Figure \ref{fig:bertviz} shows BertViz's head view method on a small input sequence with similar language. In \ref{fig:bertviz-full}, the full connections are displayed with opacity representing the relative weight applied in all heads of the first encoder layer. In \ref{fig:bertviz-isolated}, a subword of ``unresponsive" is highlighted, and inbound self-attention weights are indicated from the other tokens in the sequence.

This view is perhaps more compelling for language modeling use-cases (e.g., masked language modeling), which focus more on the learned representations of each token. It is unclear whether these attention weights provide much practical value for gaining true understanding of a Transformer-based classifier's behavior. 

\subsection{Evaluating the Quality of Explanations}

\begin{figure*}[th]
\centering
    \begin{subfigure}[Distribution of local Lipschitz values.]{
        \centering 
        \includegraphics[width=0.45\textwidth]{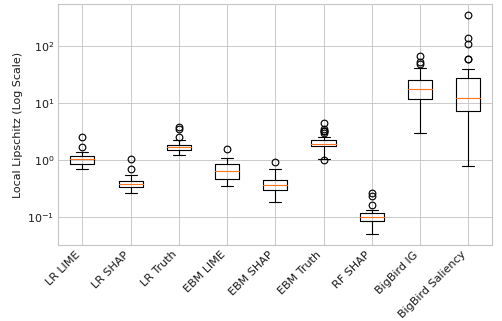}
        \label{fig:lipschitz-box}}
    \end{subfigure}%
    \begin{subfigure}[Distribution of infidelity values.]{
        \centering
        \includegraphics[width=0.45\textwidth]{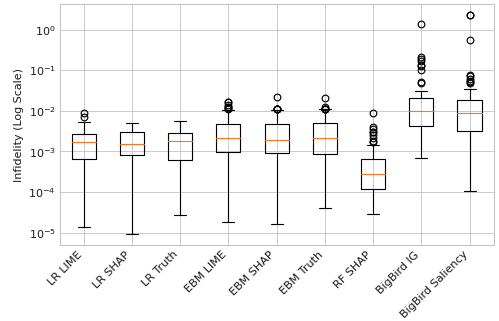}
        \label{fig:infidelity-box}}
    \end{subfigure}
    \caption{Evaluation metrics comparing models and attribution types.}
    \label{fig:boxplots}
    \vskip -0.1in
\end{figure*}

\subsubsection{Local Lipschitz}
We first explored the local Lipschitz measure \cite{alvarezmelis2018robustness} as a metric for quantifying the sensitivity of \emph{post hoc} input attribution. 

For our experiment, we used a sample of 35 documents and required a neighborhood size of 15 perturbations for each document. For each document, we calculated local Lipschitz values for each of the perturbations and retained the maximizing perturbation. The distribution of these values by model and attribution type can be seen in Figure \ref{fig:lipschitz-box}, which shows a great deal of variation between models as well as attribution methods. According to Alvarez-Melis and Jaakkola \citeyearpar{alvarezmelis2018robustness}, there is no globally optimal value for the local Lipschitz -- we can, however, use it to compare the relative sensitivity of attributions methods across models and for attributions on the same model. We can see that both attributions for BigBird have higher local Lipschitz values than all traditional methods, indicating that the available attributions are more sensitive to slight changes in the model inputs. These results show that this is still the case when using a common perturbation method. The random forest model with SHAP attributions produces the lowest local Lipschitz values, indicating that it is relatively insensitive to small input changes.

\subsubsection{(In)Fidelity}
Since there is no need to repeatedly perturb inputs, we ran our infidelity experiment on a sample of 100 documents. The results are given in Figure \ref{fig:infidelity-box}, and it is clear that there is somewhat less pronounced variation between models than with local Lipschitz. BigBird's attribution methods have higher infidelity than any of the traditional methods, indicating that the features identified by the attribution methods have less of an impact on the model's output when Gaussian noise is introduced. 

Another interesting property of infidelity can be seen in the truly interpretable models: for LR and EBM, we can compare the infidelity measures for LIME and SHAP to the infidelity for the true local explanations. In the case of EBM, the true local explanation has a slightly higher infidelity than both LIME and SHAP; for LR, the true explanations have an infidelity that lies between LIME and SHAP. This is a somewhat surprising and counterintuitive result, which suggests that future work should explore better ways to apply infidelity to NLP problems.

\subsubsection{Examining the Tradeoff between Explanation Quality and Model Performance}
With a common framework in place to evaluate the behavior of available explanations, we can now examine the trade-off between explainability and model performance. Table \ref{quantifying-explainability} illustrates this comparison for the models and explanations explored in the previous sections. As shown previously, it is clear that BigBird and LR have the highest AUC values; we can now also quantitatively compare their interpretability features through the quality metrics. In summary, LR has worse performance with substantially better explanations (in addition to the fact that it is fully interpretable). As in any practical setting, the ultimate decision will require weighing the relative importance of incremental model improvement against the fidelity of available explanations. 

Multi-objective optimization is one potential approach for such a weighting scheme in a practical setting: practitioners could weight the relative importance of raw model performance (F-score, precision, recall, AUC, etc.) against any number of explanation quality metrics to identify the model which best balances the Pareto frontier between interpretability/explainability and performance. This framework would also allow users to place constraints on certain metrics (e.g. requiring a minimum precision of 80\%) as well as incorporating any other desired quality of the final model (e.g. model size or number of parameters). 

\begin{table}[t]
\caption{AUC, Infidelity, and Lipschitz values for all tested algorithms and attribution methods.}
\label{quantifying-explainability}
\vskip -0.3in
\begin{center}
\begin{small}
\begin{sc}
\begin{tabular}{lccc}
\toprule 
Model & AUC & Infidelity & Lipschitz \\
\midrule
EBM LIME         & 76.50  & 0.003482 & 0.674207 \\
EBM SHAP         & 76.50  & 0.003325 & 0.404805 \\
EBM Truth        & 76.50  & 0.003591 & 2.097308 \\
BigBird IG       & 83.59  & 0.034362 & 21.500845 \\
BigBird Saliency & 83.59  & 0.064262 & 32.830019 \\
LR LIME          & 81.07  & 0.001947 & 1.07602 \\
LR SHAP          & 81.07  & 0.001838 & 0.41027 \\
LR Truth         & 81.07  & 0.001918 & 1.838809 \\
RF SHAP          & 79.30  & 0.000613 & 0.105773 \\
\bottomrule
\end{tabular}
\end{sc}
\end{small}
\end{center}
\vskip -0.15in
\end{table}

\section{Conclusions}
In this work, we explored practical aspects of interpretable and explainable ML methods within a case study of clinical text classification. We demonstrated some of the interpretability tools available to healthcare NLP practitioners, discussed existing definitions for explainability and interpretability, and we introduced a framework which can evaluate the quality of explanations across text classification models, including the infidelity and relative sensitivity of the presented attributions. We showed that this evaluation framework can be used in conjunction with preferred model evaluation metrics to provide a Pareto frontier illustrating the trade-off between a model's test set performance and the quality of the explanation methods available to it. This can be a very useful tool for practitioners in many domains, including healthcare, in which the data scientist is tasked with building and identifying an optimal model based on the desired level of human interpretability, which varies from problem to problem. It will also be useful for developers of AutoML solutions -- the introduction of quantitative metrics for explanation quality presents an opportunity to jointly optimize for such metrics in addition to raw model performance.

We showed that the interpretable ML techniques which are not typically used for text classification perform suboptimally despite their interpretable properties. In some cases, a deterioration in model quality may be acceptable for the purpose of full model interpretability, although other methods for producing sparse models (e.g. LASSO regression) may provide better performance. Through direct comparison between LIME, SHAP, and true local explanations, we show that LIME falls short on such high-dimensional tasks, where sparse attributions do not necessarily reflect the true underlying behavior of the model. We also demonstrated current work in visualizing the underlying behavior of BERT-style models, which attempts to build progress toward a sense of interpretability for these black-box models.

\subsection{Future Work}
Future work in both explainable and interpretable ML will undoubtedly be a crucial factor in the effort to accelerate adoption within the healthcare sector. These efforts should include additional development across the spectrum of interpretability in addition to the goal of ever-improving model performance. Progress in this area will be aided by the development and use of common evaluation metrics, which seek to quantify the quality of explanations provided for black-box models. The essence of this effort is to NLP practitioners in answering the question, ``\emph{How much} better or worse are the SHAP attributions for a random forest compared to saliency for a BERT classifier?" We make progress in this direction by introducing and open-sourcing prototypical methods for preparation-agnostic document perturbation within local neighborhoods, but additional exploration and improvement will undoubtedly aid adoption.

In addition to better understanding of explanation quality, future work in reducing the computational complexity of purely interpretable and provably optimal ML solutions will allow for more competitive use in large-scale text classification.

\section*{Acknowledgements}
The authors would like to thank the organizers of the Interpretable ML in Healthcare workshop and ICML, as well as the anonymous reviewers for their feedback and advice. We also acknowledge Digital Reasoning / Smarsh, Inc. and our colleagues for their support.

\newpage

\bibliography{refs}
\bibliographystyle{icml2021}

\end{document}